\documentclass[pdflatex]{nolta}
\usepackage[fleqn]{amsmath}
\usepackage{newtxtext}
\usepackage[varg]{newtxmath}
\usepackage{bm}
\usepackage{mathtools}
\usepackage{booktabs}
\usepackage{makecell}

\Vol{2}%
\No{3}%
\Year{2011}%
\Month{10}%

\title{Quantitative attractor analysis of high-capacity kernel Hopfield networks}

\AUTHOR*{%
\author{Akira Tamamori}{1}\orcid{0009-0000-8893-0058}
}

\AFFILIATE{ \affiliate{Faculty of Information Science, Aichi Institute
    of Technology\\1247 Yachigusa, Yakusa-cho, Toyota-shi, Aichi
    470-0392, Japan}{1} }

\received{10}{27}{20XX}
\revised{12}{29}{20XX}
\published{7}{1}{20XX}

\begin{document}

\begin{abstract}
  Kernel-based learning methods such as Kernel Logistic Regression
  (KLR) can substantially increase the storage capacity of Hopfield
  networks, but the principles governing their performance and
  stability remain largely uncharacterized. This paper presents a
  comprehensive quantitative analysis of the attractor landscape in
  KLR-trained networks to establish a solid foundation for their
  design and application. Through extensive, statistically validated
  simulations, we address critical questions of generality,
  scalability, and robustness. Our comparative analysis shows that KLR
  and Kernel Ridge Regression (KRR) exhibit similarly high storage
  capacities and clean attractor landscapes under typical operating
  conditions, suggesting that this behavior is a general property of
  kernel regression methods, although KRR is computationally much
  faster. We identify a non-trivial, scale-dependent law for the
  kernel width $\gamma$, demonstrating that optimal capacity requires
  $\gamma$ to be scaled such that $\gamma N$ increases with network
  size $N$. This finding implies that larger networks require more
  localized kernels, in which each pattern's influence is more
  spatially confined, to mitigate inter-pattern interference. Under
  this optimized scaling, we provide clear evidence that storage
  capacity scales linearly with network size~($P \propto
  N$). Furthermore, our sensitivity analysis shows that performance is
  remarkably robust with respect to the choice of the regularization
  parameter $\lambda$. Collectively, these findings provide a concise
  set of empirical principles for designing high-capacity and robust
  associative memories and clarify the mechanisms that enable kernel
  methods to overcome the classical limitations of Hopfield-type
  models.
\end{abstract}

\begin{keywords}
  Hopfield network, kernel logistic regression, associative memory,
  storage capacity, noise robustness, attractor analysis
\end{keywords}

\maketitle

\section{Introduction}
Associative memory (AM) is a fundamental cognitive function that
enables the retrieval of stored information based on partial or noisy
cues. Artificial neural networks, particularly the Hopfield
network~\cite{Hopfield1982}, have long served as foundational models
for AM due to their biologically plausible recurrent dynamics and the
interpretation of stored memories as stable states (attractors) of an
energy function. The recall process in a Hopfield network can be
viewed as the network state evolving toward a local minimum of its
energy landscape, corresponding to the stored pattern that is most
similar to the input cue.

Despite their conceptual elegance and biological grounding,
traditional Hopfield networks trained with the Hebbian learning
rule~\cite{Hopfield1982} exhibit severe limitations. Their storage
capacity is theoretically bounded at a pattern-to-neuron ratio $(P/N)$
of approximately 0.14~\cite{Amit1985}. Exceeding this limit leads to
catastrophic interference, recall errors, and the proliferation of
\textit{spurious attractors}, which do not correspond to any learned
pattern yet still act as fixed points of the network
dynamics~\cite{Amit1985b}. These spurious attractors compete with
desired memories, reduce their basins of attraction, and significantly
impair noise robustness.

To overcome these limitations, several approaches have reframed
associative memory learning as a supervised optimization problem in
which synaptic weights are tuned to make desired patterns stable fixed
points. Linear methods, such as the pseudoinverse
rule~\cite{Kohonen1989} and Linear Logistic Regression
(LLR)~\cite{MacKay2002}, provide moderate improvements in storage
capacity and robustness. However, their performance remains
fundamentally constrained by the requirement of linear separability,
either in the input space or in the neuron’s input--output mapping.

Nonlinear kernel methods offer a powerful means to transcend these
constraints. Techniques such as Support Vector Machines (SVMs), Kernel
Ridge Regression (KRR), and Kernel Logistic Regression (KLR)
implicitly map data into high- or even infinite-dimensional feature
spaces, where complex nonlinear relationships become linearly
separable~\cite{Scholkopf2001}. Applying this framework to associative
memory learning holds the potential to dramatically increase storage
capacity and enhance noise robustness.

Our previous work~\cite{tamamori2025} revealed the remarkable
potential of KLR for training high-capacity Hopfield networks,
demonstrating storage capacities far exceeding the classical
limit. This initial discovery, however, left fundamental questions
regarding the generality, scalability, and robustness of this approach
unanswered. For instance, is the choice of the loss function critical?
How does performance scale with network size $N$? And which
hyperparameters govern the stability?

This paper aims to answer these questions through a comprehensive
quantitative analysis. We systematically investigate the attractor
landscape by conducting a series of rigorous, statistically validated
experiments, including (1) a direct comparison with KRR, (2) a
multi-scale analysis of storage capacity, and (3) a detailed
hyperparameter sensitivity study.

The main contributions of this work are as follows:
\begin{enumerate}
\item We confirm and statistically validate that KLR-trained networks
  achieve significantly higher storage capacity and noise robustness,
  exhibiting a remarkably clean attractor landscape, nearly devoid of
  spurious states under typical noise conditions.
  % with near-zero rates of spurious fixed points.
\item We reveal that KLR and KRR exhibit qualitatively similar recall
  performance, suggesting that the high-capacity nature is a general
  property of kernel regression methods in this context. However, KRR
  holds a significant advantage in computational speed.
\item We uncover a non-trivial scaling law for the kernel width
  $\gamma$, demonstrating that optimal capacity is achieved when
  $\gamma$ is scaled appropriately with $N$. Under these optimized
  conditions, we confirm that the storage capacity $P$ scales linearly
  with $N$.
\item We show that the network's high performance is robust across a
  reasonable range of hyperparameter values.
\end{enumerate}
These findings offer deep insights into how kernel-based learning
reshapes the dynamical system of a Hopfield network to create a highly
effective associative memory. By moving beyond the initial discovery
of high performance to a deeper characterization of its generality,
scaling properties, and robustness, this work establishes a solid
foundation for the design and application of high-capacity,
Hopfield-type models.

The remainder of this paper is organized as
follows. Section~\ref{sec:related_work} reviews related work,
positioning our study within the context of both classical Hopfield
networks and modern kernel-based
approaches. Section~\ref{sec:model_methods} details the KLR and KRR
network models, introduces our geometric stability metric ``Pinnacle
Sharpness,'' and describes the experimental
methodology. Section~\ref{sec:results} presents the results of our
comprehensive simulations, including the attractor landscape analysis,
comparative study with KRR, scaling analysis, and hyperparameter
sensitivity analysis. Section~\ref{sec:discussion} discusses the
implications of our findings, focusing on the mechanism of
self-organization and practical design principles. Finally,
Section~\ref{sec:conclusion} concludes the paper with a summary of our
contributions and directions for future research.

\section{Related Work}
\label{sec:related_work}
Traditional approaches to Hopfield network learning, primarily relying
on Hebbian rules, suffer from limited storage
capacity~\cite{Hopfield1982, Amit1985} and the emergence of spurious
attractors~\cite{Amit1985b}. Linear learning methods, such as the
pseudoinverse rule~\cite{Kohonen1989} and those based on Linear
Logistic Regression (LLR)~\cite{MacKay2002}, offer some improvement
but remain fundamentally constrained by the linear separability of
stored patterns in the input space.

The application of nonlinear kernel methods to associative memory
learning represents a promising direction. Kernel Associative Memories
(KAM)~\cite{Nowicki2010} first proposed leveraging high-dimensional
feature spaces to theoretically achieve vast storage capacities.
Early theoretical analyses, such as those by Caputo and
Niemann~\cite{Caputo2002} using spin-glass theory, provided initial
evidence that polynomial and Gaussian kernels could indeed yield a
notable improvement over the classical Hopfield model.  Saltz and
Belanche~\cite{Saltz2016} proposed a kernelized Bidirectional
Associative Memory (BAM) by directly applying the kernel trick to the
Hebbian energy function. They demonstrated improved capacity and noise
tolerance using various inference methods such as hill
climbing. However, their approach primarily relied on Hebbian-like
fixed weights and did not optimize the dual variables via supervised
learning, which limits the potential storage capacity compared to
discriminative training methods.

More recently, the connection between Hopfield-type networks and
modern deep learning architectures has revitalized the field. A
landmark paper by Ramsauer et al.~\cite{ramsauer2021} demonstrated
that the attention mechanism in Transformers can be interpreted as a
modern Hopfield network with a continuous state representation and an
updated energy function. These Modern Hopfield Networks (MHNs) have
shown immense storage capacities, scaling polynomially or even
exponentially with the feature dimension. This work firmly positions
the study of Hopfield-like energy-based models at the heart of current
AI research.

While these important lines of research, ranging from classical KAM to
MHNs, have demonstrated the potential for high-capacity associative
memory, their focus has primarily been on the static properties of the
energy function or on performance in specific pattern reconstruction
tasks. A detailed and quantitative understanding of the attractor
landscape and recall dynamics that arise from practical supervised
kernel methods such as KLR and KRR remains largely unexplored. Key
questions persist: What happens to spurious attractors in these
high-capacity regimes? What are the primary failure modes of recall?
What are the geometric properties of the attractors themselves?

The present study aims to address this gap. We connect the high-level
capacity analysis of KAM and MHN with the classical dynamical systems
perspective on Hopfield networks. By conducting a detailed
quantitative analysis of the attractor landscape formed by KLR and
comparing it with KRR, for which we have separately demonstrated
strong performance and computational advantages~\cite{tamamori2025b},
we provide a microscopic and dynamical explanation of how these
kernel-based methods achieve their notable performance, thereby
complementing the existing body of research. A systematic empirical
comparison with other high-capacity associative memory models,
including modern Hopfield networks and related variants, remains an
important direction for future work.

\section{Model and Methods}
\label{sec:model_methods}
\subsection{Network Model}
We consider a standard Hopfield network composed of $N$ fully
connected bipolar neurons, with states $s_i \in \{-1, 1\}$ for
$i = 1, \dots, N$.  The network evolves over discrete time steps.
Given a state $\bm{s}(t)$ at time $t$, the next state $\bm{s}(t+1)$ is
determined by updating each neuron based on its activation potential.

\subsection{Learning Algorithms for Comparison}
For comparison, we briefly describe the conventional Hebbian learning
rule and Linear Logistic Regression (LLR), as implemented in our
previous work~\cite{tamamori2025}.

\subsubsection{Hebbian Learning}
The Hebbian learning rule defines the synaptic weights $W_{ij}$ based
on correlations between the activities of neurons $i$ and $j$ across
$P$ stored patterns $\{\boldsymbol{\xi}^\mu\}_{\mu=1}^P$:
\begin{equation}
W_{ij} = \frac{1}{N} \sum_{\mu=1}^P \xi_i^\mu \xi_j^\mu, \quad (i \neq j), \quad W_{ii} = 0.
\end{equation}
Recall is performed using standard linear threshold dynamics:
\begin{equation}
s_i(t+1) = \text{sign}\left(\sum_{j \neq i} W_{ij} s_j(t)\right),
\end{equation}
where $\text{sign}(z) = 1$ if $z \geq 0$ and $-1$ if $z < 0$.

\subsubsection{Linear Logistic Regression (LLR)}
In the LLR approach~\cite{MacKay2002}, learning for each neuron $i$ is
formulated as predicting $\xi_i^\mu$ from the remaining neuron states
$\xi_{j \neq i}^\mu$.  A linear logistic regression model is trained
for each neuron to obtain a weight vector $\bm{w}_i$.  The logit for
neuron $i$ given a network state $\bm{s}$ is computed as
$h_i(\bm{s}) = \sum_{j \neq i} W_{ij} s_j$.  The weights are optimized
by minimizing a regularized negative log-likelihood using gradient
descent.  After training, the resulting $N \times N$ symmetric weight
matrix $W_{ij}$ is used for recall with the same dynamics as Hebbian
learning.

\subsubsection{Kernel Logistic Regression (KLR) Hopfield Network}
The primary focus of this work is the Hopfield network trained using
Kernel Logistic Regression (KLR), as proposed in our previous
study~\cite{tamamori2025}.  KLR treats the learning for each neuron
$i$ as an independent binary classification task, predicting the target state
$t_i^\mu = (\xi_i^\mu + 1)/2 \in \{0, 1\}$ from the input pattern
$\boldsymbol{\xi}^\mu \in \{-1, 1\}^N$.

KLR employs the kernel trick~\cite{Scholkopf2001} to implicitly map
input patterns into a high-dimensional feature space.  The logit for
neuron $i$ given a pattern $\boldsymbol{\xi}$ is computed as a
weighted sum of kernel similarities to the stored patterns:
\begin{equation}
  h_i(\boldsymbol{\xi}) = \sum_{\mu=1}^P \alpha_{\mu i} K(\boldsymbol{\xi}, \boldsymbol{\xi}^\mu).
  \label{eq:klr_potential}
\end{equation}
We adopt the Radial Basis Function (RBF) kernel
$K(\bm{x}, \bm{y}) = \exp(-\gamma \|\bm{x} - \bm{y}\|^2)$, which
measures pairwise similarity between patterns in the feature
space. The dual variables
$\boldsymbol{\alpha}_i = [\alpha_{1i}, \dots, \alpha_{Pi}]^T$ for each
neuron $i$ are learned by minimizing the L2-regularized negative
log-likelihood objective function. The objective function (loss
function) $L_i$ to be minimized for neuron $i$ is given by:
\begin{equation}
  L_i(\boldsymbol{\alpha}_i) = - \sum_{\mu=1}^{P} \left[ t^{\mu}_i \log(\sigma(h_i(\boldsymbol{\xi}^\mu))) + (1 - t^{\mu}_i) \log(1 - \sigma(h_i(\boldsymbol{\xi}^\mu))) \right] + \frac{\lambda}{2}\boldsymbol{\alpha}_i^{\top} \bm{K}\boldsymbol{\alpha}_i,
\end{equation}
$h_i(\boldsymbol{\xi}^\mu)$ is the logit from
Eq.~(\ref{eq:klr_potential}), $\sigma(z) = 1/(1+e^{-z})$ is the
sigmoid function, and $\lambda$ is the L2-regularization
parameter. This optimization is performed via gradient descent.

\subsubsection{Kernel Ridge Regression (KRR) Hopfield Network}
As a key point of comparison, we also investigate networks trained
with Kernel Ridge Regression (KRR)~\cite{tamamori2025b}.  Like KLR,
KRR is a kernel-based method that learns a set of dual variables,
$\boldsymbol{\alpha} \in \mathbb{R}^{P\times N}$. However, it differs
from KLR in two fundamental aspects: the task formulation and the
optimization procedure.

First, KRR frames the learning problem as a regression task rather
than a classification task. The goal is to make the network's output
$h_i(\cdot)$ directly approximate the target bipolar states
$\{-1, 1\}^N$ by minimizing a squared error loss. The objective
function for each neuron $i$ is to minimize:
\begin{equation}
L_i(\boldsymbol{\alpha}_i)= \sum_{\mu = 1}^{P} (\xi^{\mu}_{i} - h_i(\boldsymbol{\xi}^{\mu}))^{2} + \frac{\lambda}{2}\boldsymbol{\alpha}_i^{\top} \bm{K}\boldsymbol{\alpha}_i.
\end{equation}

Second, and most critically from a practical standpoint, this
squared-error objective function leads to a closed-form, non-iterative
solution for the dual variables $\boldsymbol{\alpha}$. The optimal
$\boldsymbol{\alpha}$ can be found directly by solving the following
system of linear equations:
\begin{equation}
(\bm{K} + \lambda \bm{I}) \boldsymbol{\alpha} = \bm{Y}, 
\end{equation}
where $\bm{K}$ is the $P \times P$ kernel (Gram) matrix
$K_{\mu \nu} = K(\boldsymbol{\xi}^{\mu}, \boldsymbol{\xi}^{\nu})$,
$\bm{I}$ is the identity matrix, and $\bm{Y}$ is the $P \times N$
matrix of stored patterns. This non-iterative solution makes the
training process of KRR significantly faster than the gradient-based
optimization required for KLR, a difference we quantify in our
experiments.

Once $\boldsymbol{\alpha}$ is learned, the recall dynamics are
identical to those of the KLR network, as described in Section 3.4.

\subsection{Lyapunov Function Candidate for Landscape Analysis}
To provide a theoretical framework for analyzing the attractor
landscape, we define a Lyapunov function candidate for the KLR network
dynamics. While the synchronous update rule of the KLR network does
not guarantee the monotonic decrease of a traditional energy function,
we can define a scalar function $V(\bm{s})$ whose landscape geometry
reflects the restorative forces of the system. We define this function
as:
\begin{equation}
  V(\bm{s}) = - \sum_{k=1}^{N} s_k h_k(\bm{s}),
\end{equation}
where $h_k(\bm{s})$ is the input potential to neuron $k$ as defined in
Eq.~(\ref{eq:klr_potential}). This function measures the total
alignment between the neuron states $s_k$ and their corresponding
input potentials $h_k(\bm{s})$. A lower value of $V(\bm{s})$ indicates
a state that is more consistent with the dynamics of the network;
thus, the network's dynamics can be viewed as a process that seeks to
find local minima of this function $V(\bm{s})$.

The local stability of an attractor $\boldsymbol{\xi}^{\mu}$ can then
be characterized by the steepness of the landscape of $V(\bm{s})$ in
its vicinity. As will be detailed in our experimental analysis, the
gradient of this function, $\nabla V(\bm{s})$, provides a powerful
tool for quantitatively characterizing the properties of the attractor
landscape. The primary scope of this paper is to use this tool for an
empirical and phenomenological analysis of the landscape. A detailed
theoretical investigation into the mathematical structure of
$V(\bm{s})$ and its resulting phase diagram is the subject of our
companion paper~\cite{tamamori2025c}. Therefore, this paper does not
delve into the rigorous theoretical proofs of stability, but rather
establishes the empirical foundations upon which such theories can be
built.

\subsection{Recall Process}
The recall process in KLR-trained networks differs from that in
Hebbian and LLR-based networks, as it does not rely on an explicit
$N \times N$ weight matrix.  Given a network state $\bm{s}(t)$ at time
$t$, the next state $\bm{s}(t+1)$ is determined by the sign of each
neuron's activation potential relative to a threshold
$\boldsymbol{\theta}$:
\begin{equation}
s_i(t+1) = \text{sign}(h_i(s(t)) - \theta_i),
\end{equation}
where 
\begin{equation}
h_i(s(t)) = \sum_{\mu=1}^P \alpha_{\mu i} K(s(t), \xi^\mu)
\end{equation}
is the logit for neuron $i$.  In this study, the threshold vector was
set to $\boldsymbol{\theta} = \mathbf{0}$, i.e., $\theta_i = 0$ for
all $i$.  The recall process involves computing kernel values between
the current state and all $P$ stored patterns at each iteration.

\subsection{Experimental Setup}
We conducted a series of large-scale numerical simulations to
systematically analyze the attractor landscape and performance
characteristics of kernel-based Hopfield networks. All simulations
were implemented in Python 3.13 using the \verb+NumPy+ 2.1.3 and
\verb+SciPy+ 1.15.2 libraries and were executed on a standard
workstation equipped with an Intel Core i9-9900K CPU and 64 GB of
RAM. No GPU acceleration was used.

\subsubsection{Network and Learning Parameters}
Unless otherwise specified, the network size was set to $N=500$ for
detailed attractor analysis and $N=100$ for comparative and
sensitivity analyses. Stored patterns $\boldsymbol{\xi}^\mu$ were
generated as random bipolar vectors $\{-1, 1\}^N$, with each element
sampled independently with equal probability.

For the KLR learning process, we used gradient descent with a learning
rate $\beta = 0.1$ and number of parameter updates $M = 200$. For the
KRR learning process, the closed-form solution was used. The
L2-regularization parameter $\lambda$ was set to a default value of
$\lambda = 0.01$ for most experiments, except for the sensitivity
analysis. As shown in Appendix~\ref{sec:appendix_convergence}, the
learning curve for a challenging high-load condition ($N=500$,
$P/N=4.0$) demonstrates that the loss substantially decreases within
the initial $\sim$150 steps and approaches a plateau, confirming that
200 updates are adequate to achieve a well-converged model.

The RBF kernel width $\gamma$ is a critical parameter.  A common
heuristic for kernel methods is to scale the width inversely with the
input dimension, such as $\gamma = 1/N$, to maintain a constant level
of pattern similarity irrespective of
$N$~\cite{Scholkopf2001}. However, our preliminary analysis revealed
that the optimal $\gamma$ scales with $N$ in a non-trivial manner. To
determine these optimal, $N$-dependent values, we performed a
simplified grid search. For each network size $N$ (100, 250, 500), we
tested several values of the scaling factor $c$ in $\gamma = c/N$,
where $c$ ranged from 0.5 to 10.0. We evaluated the average Target
Recall Rate across multiple high-load conditions ($P/N$ from 1.5 to
4.0) with a fixed initial similarity of 1.0. This search revealed that
the optimal factor $c_{\text{opt}}$ increases with $N$ (e.g.,
$c_{\text{opt}} \approx 2$ for $N \leq 250$ and
$c_{\text{opt}} \approx 5$ for $N = 500$), indicating that larger
networks require more localized kernels than the conventional
heuristic suggests.  Therefore, for the main experiments, we used
these empirically determined optimal $\gamma$ values to ensure a fair
comparison across scales. The detailed results of this sensitivity are
presented in Section 4.4.

\subsubsection{Simulation Experiments}
We designed four main sets of experiments to address the key questions
of this study:
\begin{enumerate}
\item \textbf{Attractor Landscape Analysis} ($N=500$): To characterize
  the detailed structure of the attractors, we evaluated recall
  dynamics across a wide range of storage loads ($P/N$ from 0.05 to 4.0)
  and initial similarities (from 0.05 to 1.0).
\item \textbf{Comparative Analysis} (KLR vs. KRR, $N=100$): To compare
  KLR with KRR, we measured their recall performance and training time
  under identical conditions, sweeping $P/N$ from 0.1 to 6.0.
\item \textbf{Scaling Analysis}: To verify the capacity scaling law,
  we performed the capacity analysis for three different network sizes
  ($N = 100, 250, 500$) using their respective optimal gamma values.
\item \textbf{Hyperparameter Sensitivity Analysis} ($N=100$): To
  assess the robustness of the model, we evaluated the recall
  performance while systematically varying the parameters $\gamma$
  (via $c$) and $\lambda$ at fixed medium ($P/N=1.5$) and high
  ($P/N=3.0$) storage loads.
\end{enumerate}

\subsection{Attractor Analysis Methodology}
\subsubsection{Recall Simulation and Convergence Criteria}
To ensure statistical reproducibility, all experiments were repeated
across five different master random seeds. Each master seed determined
the generation of stored patterns and the initialization of the
learning process. Within each master seed trial, further randomization
for initial state generation was controlled by separate seeds to
ensure diverse sampling.

For each combination of $P/N$ and initials similarity, five noisy
initial states were generated for each of the $P$ stored patterns,
resulting in a total of $5 \times P$ recall trials per condition per
master seed.

The recall dynamics were simulated for a maximum of 30 discrete,
synchronous update steps. Convergence was classified into three
categories:

\begin{itemize}
\item \textit{Fixed Point}: The state remains unchanged for one
  consecutive step.
\item \textit{Limit Cycle}: The state returns to any previously
  visited state within the trial.
\item \textit{Not Converged}: Neither of the above occurs within the
  maximum steps.
\end{itemize}

\subsubsection{Classification of Final States}
Once a trial converges, we further analyze the nature of the final
state $\bm{s}_{\text{final}}$ to characterize the attractor landscape.
Each final state $\bm{s}_{\text{final}}$ reached in a recall trial was
classified into one of the following categories based on strict
equality checks against the $P$ stored patterns:

\begin{itemize}
\item \textit{Target Pattern}: $\bm{s}_{\text{final}}$ is
  identical to the original target pattern $\boldsymbol{\xi}^{\nu}$
  from which the initial state was generated.
\item \textit{Other Learned Pattern}: $\bm{s}_{\text{final}}$ is
  identical to one of the other $P-1$ learned patterns
  ($\boldsymbol{\xi}^{\nu}$ where $\mu \neq \nu$).
\item \textit{Spurious Attractor}: $\bm{s}_{\text{final}}$ is a fixed
  point or a limit cycle state that is not identical to any of the $P$
  learned patterns.
\end{itemize}

In this work, a fixed point is classified as spurious unless it
exactly matches one of the stored patterns. This criterion is stricter
than the classical definition, which often permits small
Hamming-distance deviations. We adopt this strict definition to
eliminate ambiguity in the quantitative analysis of attractor
statistics.

The validity of this strict, equality-based classification is
empirically supported by the distribution of Hamming distances between
final states and their nearest learned patterns. As detailed in
Appendix~\ref{sec:appendix_Attractor_Classification}, this analysis
confirms a clear separation, with no spurious attractors found in
close proximity to any learned pattern.

\subsubsection{Quantitative Metrics}

Based on the above classifications, we computed the following
quantitative metrics, which are averaged over all trials and all
master seeds for each experimental condition:

\begin{itemize}
\item \textit{Target Recall Rate}: The proportion of trials classified
  as \textit{Target Pattern}. This metric quantifies the effective size
  of an attractor's basin of attraction.
\item \textit{Other Learned Recall Rate}: The proportion of trials
  classified as \textit{Other Learned Pattern}. This measures the rate
  of ``mis-recall.''
\item \textit{Spurious Fixed Point Rate}: The proportion of trials
  classified as \textit{Spurious Attractor} that converged to a fixed
  point.
\item \textit{Cycle Rate}: The proportion of trials that converged to
  a limit cycle with a period of 2 or more.
\item \textit{Not Converged Rate}: The proportion of trials that did
  not converge to either a fixed point or a limit cycle within the
  maximum number of steps.
\item \textit{Fixed Point Rate}: The proportion of trials that
  converged to any fixed point (including target, other learned, and
  spurious). This is equivalent to
  1.0 $-$ (\textit{Cycle Rate} + \textit{Not Converged Rate}).
\item \textit{Average Steps to Converge}: The average number of update
  steps required to reach a fixed point or detect a cycle, averaged
  only over the trials that converged.
\end{itemize}

\section{Results}
\label{sec:results}
In this section, we present the results of our extensive numerical
simulations. All plotted data points represent the mean values
averaged across five master random seeds, and the shaded regions or
error bars indicate the 95\% confidence intervals, ensuring
statistical robustness of our findings.

A foundational aspect of the network's dynamics is its convergence
behavior. While a formal proof of convergence for KLR-trained Hopfield
networks with synchronous updates is not established, our experiments
revealed a remarkable degree of stability. Across the tens of
thousands of recall trials conducted for this study, every single
trial converged to a stable state (either a fixed point or a limit
cycle) within the maximum of 30 update steps.

Furthermore, a detailed analysis of non-stationary dynamics showed
that the rate of convergence to limit cycles (\textit{Cycle Rate}) was
negligible ($< 0.1\%$) across all tested parameter regimes (see
Appendix~\ref{sec:appendix_dynamics} for a representative plot). This
empirical observation that the dynamics are overwhelmingly dominated
by convergence to fixed points validates our subsequent focus on the
analysis of these fixed point attractors and their properties.

\subsection{Performance and Attractor Landscape Characteristics ($N=500$)}

\begin{figure}[t]
  \begin{center}
    \includegraphics[width=0.99\hsize]{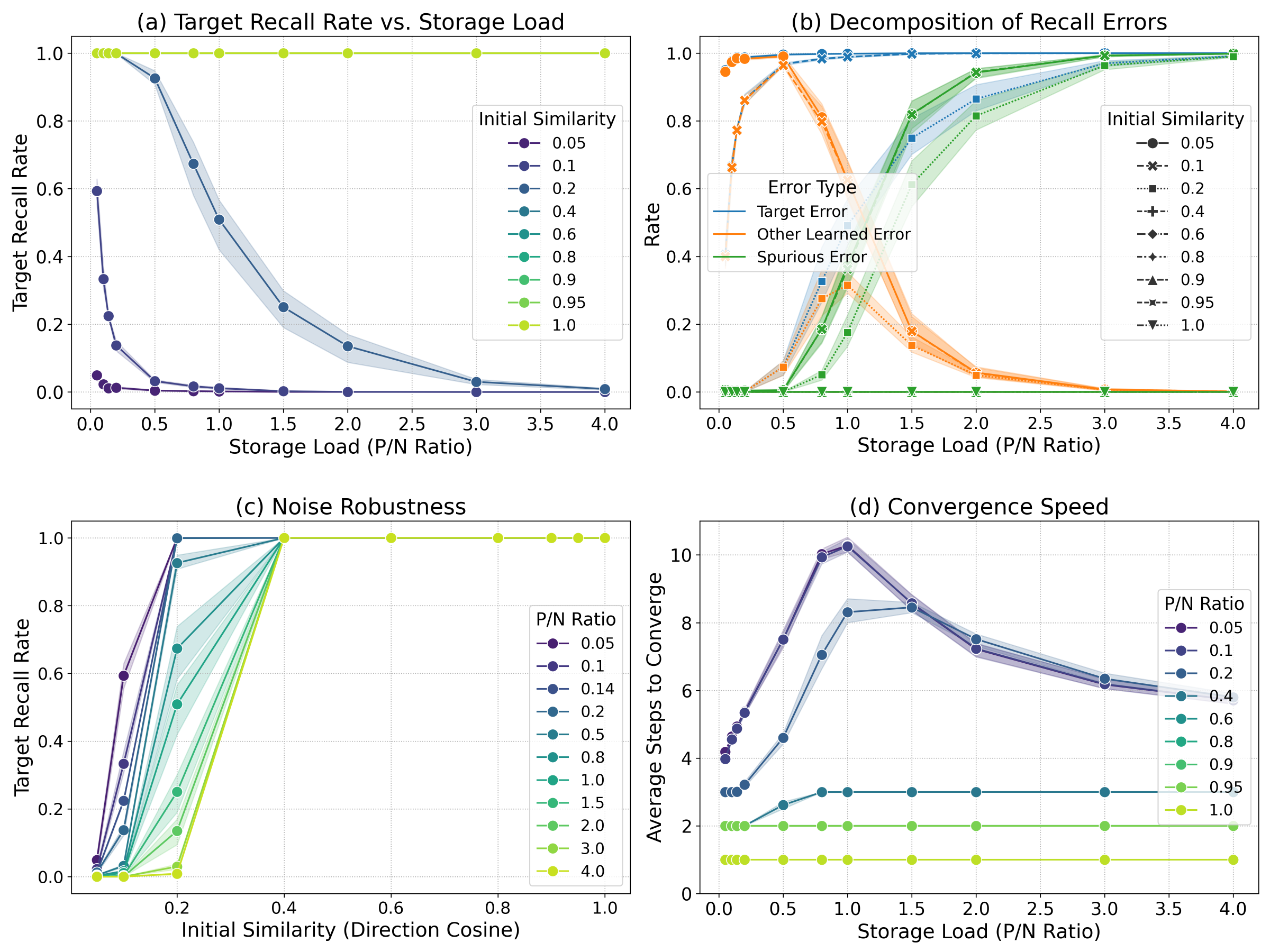}
\end{center}
\caption{ Performance and attractor landscape characteristics of the
  KLR-trained network ($N=500$).  All curves show the mean values
  averaged across five master random seeds, with shaded regions
  representing the 95\% confidence intervals.  (a) \textbf{Capacity
    Curve:} \textit{Target Recall Rate} as a function of storage load
  ($P/N$) for various initial similarities. The network maintains
  near-perfect recall up to $P/N=4.0$ for low-noise inputs.  (b)
  \textbf{Decomposition of Recall Errors:} Recall Errors (1.0 -
  \textit{Target Recall Rate}) are decomposed into convergence to
  other learned patterns and spurious attractors. Colors distinguish
  the error type, while line styles represent different initial
  similarities. For moderate noise levels, the rate of spurious errors
  is negligible, demonstrating a remarkably clean landscape where
  mis-recall to other learned patterns is the dominant failure
  mode. However, under extreme conditions (high load and high noise),
  spurious errors become significant.  (c) \textbf{Noise Robustness:}
  Target Recall Rate as a function of initial similarity for various
  storage loads ($P/N$). At low loads, recall is robust even from
  highly corrupted states.  (d) \textbf{Convergence Speed:} Average
  number of steps to convergence for various initial
  similarities. Recall is exceptionally fast, typically within 1-2
  steps for high-similarity inputs and well under 10 steps even for
  noisy, high-load conditions.}
\label{fig:main_performance}
\end{figure}

First and foremost, the KLR-trained network demonstrates exceptional
storage capacity and robustness, as shown in Fig. 1(a). For initial
states with low noise (initial similarity $\geq 0.9$), the network
achieves near-perfect recall (\textit{Target Recall Rate}
$\approx 1.0$) across the entire tested range, up to a remarkable
storage load of $P/N = 4.0$. This performance vastly surpasses the
classical Hebbian limit ($\approx 0.14$). The plot also illustrates a
graceful degradation of performance as the initial noise increases;
for instance, at a moderate noise level (initial similarity $= 0.6$),
the network maintains high performance up to $P/N \approx 2.0$ before
the recall rate begins to decline. This confirms the model's dual
strengths in both capacity and robustness.

To understand the nature of the recall failures observed at high loads
and noise levels, we decomposed the errors into their constituent
sources. The results, shown in Fig. 1(b), reveal a crucial the
detailed breakdown of recall failures.At low to medium storage loads
($P/N < 1.5$), the rate of convergence to spurious attractors
(\textit{Spurious Error}, green lines) is negligible. In this regime,
the dominant source of error is the convergence to an incorrect but
valid learned pattern (\textit{Other Learned Error}, orange
lines). This ``mis-recall'' indicates that the basins of attraction
for different stored patterns are competing with each other. However,
a distinct behavior emerges at very high storage loads ($P/N > 2.0$)
under high noise conditions (initial similarity $< 0.2$). As the
storage load increases further, the rate of mis-recall to other
learned patterns decreases, while the rate of convergence to spurious
attractors sharply increases, eventually becoming the dominant failure
mode. This suggests that in the extremely overloaded regime, the
energy landscape becomes so rugged that the system frequently gets
trapped in local minima (spurious states) that do not correspond to
any stored memory, rather than being attracted to a wrong memory.  It
is important to note that for lower noise levels (e.g., initial
similarity $\geq 0.4$), the Spurious Error remains low even at high
loads, confirming the robustness of the model within a reasonable
operating range.

The network's robustness to noise is further detailed in
Fig. 1(c). This plot shows that at low storage loads (e.g.,
$P/N \leq 1.0$), the network is highly robust, capable of perfect
recall even from initial states with similarities as low as 0.4. As
the load increases, the recall process requires a higher initial
similarity, signifying a shrinkage of the basins of attraction. The
sharp, cliff-like transitions observed for each load level indicate
that the basins of attraction are well-defined.

Finally, the recall dynamics are exceptionally fast, as illustrated in
Fig. 1(d). For high-similarity initial states, the network
consistently converges in only one or two steps. Even for highly
corrupted states in the low-similarity regime, the average number of
steps to convergence remains remarkably small, typically well below 10
steps. Notably, for these noisy inputs, the convergence time
exhibits a non-monotonic ``hump'' shape, peaking at intermediate
storage loads. The subsequent decrease in convergence time at very
high loads corresponds precisely to the regime where recall failures
dominate (cf. Fig. 1(b)).

This observation suggests that the dynamics rapidly settle into nearby
stable states, which correspond to incorrect learned patterns or
spurious fixed points, rather than conducting a prolonged search for
the correct target. Such rapid convergence represents a meaningful
practical advantage of the model.

\subsection{Comparison with Kernel Ridge Regression ($N=100$)}

\begin{figure}[t]
  \begin{center}
    \includegraphics[width=0.99\hsize]{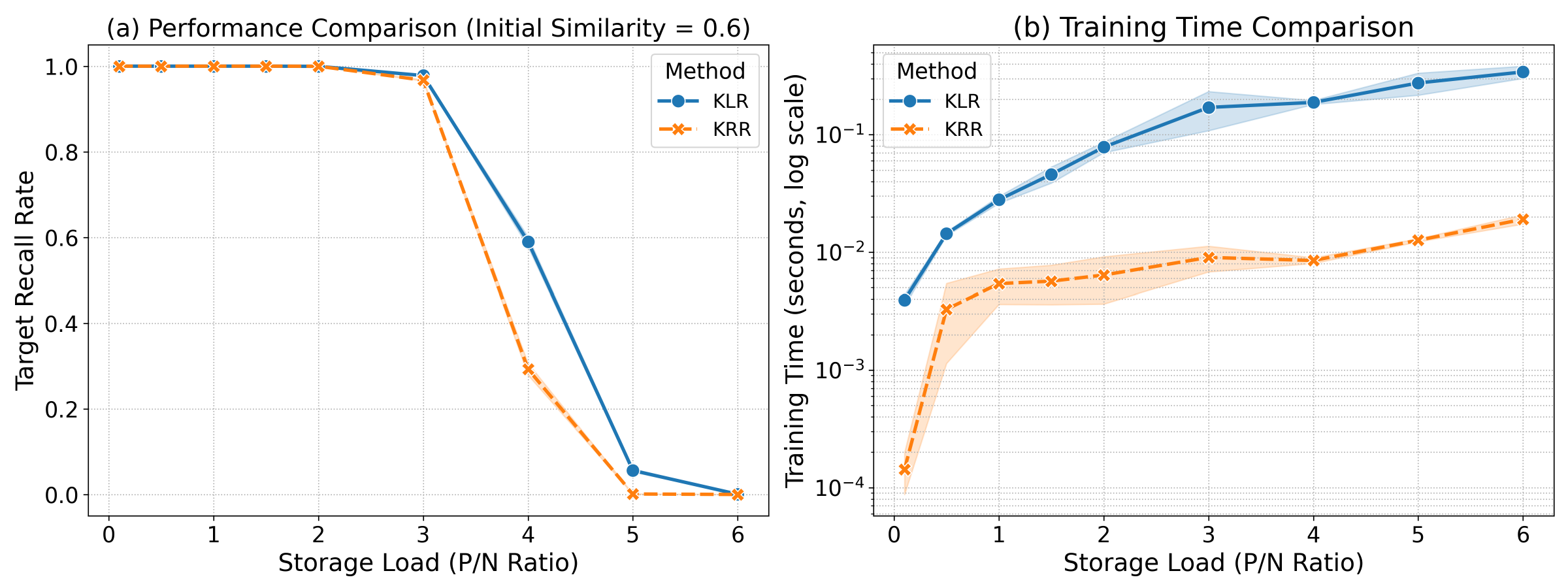}
  \end{center}
  \caption{Comparative analysis of Kernel Logistic Regression (KLR)
    and Kernel Ridge Regression (KRR) for a network with $N=100$
    neurons and an optimized kernel width ($\gamma=0.02$).  All curves
    show the mean over five master seeds, with shaded regions
    representing the 95\% confidence intervals (for performance) or
    standard deviation (for timing).  (a) \textbf{Performance
      Comparison:} \textit{Target Recall Rate} as a function of storage load
    ($P/N$) under a moderate noise condition (initial similarity =
    0.6). Both methods exhibit high and qualitatively similar
    performance, maintaining near-perfect recall up to
    $P/N \approx 3.0$.  (b) \textbf{Training Time Comparison:}
    Training time in seconds (log scale) as a function of storage
    load. The non-iterative, closed-form solution of KRR results in a
    training process that is consistently one to two orders of
    magnitude faster than the iterative gradient descent required for
    KLR.}
\label{fig:klr_vs_krr}
\end{figure}

To contextualize the properties of the KLR-trained network and address
the question of whether its high performance is a unique feature of
the logistic loss function, we conducted a direct comparative analysis
against Kernel Ridge Regression (KRR). KRR offers an alternative
kernel-based learning framework that minimizes squared error loss and,
crucially, admits a non-iterative, closed-form solution. The
comparison was performed on a network of $N=100$ neurons using the
empirically determined optimal kernel width ($\gamma = 0.02$) for this
scale.

The results of this comparison are summarized in Fig.~2. We first
evaluated the recall performance under a moderate noise condition
(initial similarity = 0.6), as shown in Fig.~2(a). The performance of
the two methods is remarkably similar. Both KLR and KRR maintain
near-perfect recall for storage loads up to $P/N \approx 3.0$. Beyond
this critical capacity, both methods exhibit a sharp yet smooth
degradation in performance, with their recall curves closely
aligned. Although minor quantitative differences exist, with KLR
displaying a slightly wider shoulder in this case, the overall
qualitative behavior is essentially identical. These findings strongly
suggest that the observed high-capacity and robust attractor dynamics
are not specific to KLR but are likely a more general property of
Hopfield-type networks trained with kernel regression methods that
effectively separate patterns in a high-dimensional feature space.

In stark contrast to their performance similarity, the two methods
exhibit a dramatic difference in computational efficiency for
training. Fig. 2(b) plots the training time as a function of storage
load. The non-iterative nature of KRR provides a substantial
advantage: its training process is consistently one to two orders of
magnitude faster than the iterative gradient descent required for KLR
across the entire range of storage loads. For instance, at
$P/N = 4.0$, KRR training completes in approximately 0.01 seconds,
whereas KLR requires over 1 second. This difference in efficiency
grows with the number of patterns, highlighting KRR as a highly
attractive and practical alternative for training large-scale
associative memories.

In summary, this comparative analysis clarifies the trade-offs between
KLR and KRR. While both methods can construct similarly
high-performance attractor landscapes, KRR achieves this with
significantly lower computational cost for training. This reinforces
the generality of our findings while also providing important
practical insights for future implementations.  It is important to
note, however, that this computational advantage applies only to the
training phase. The recall process for both KLR and KRR requires the
evaluation of kernel similarities between the current state and all
$P$ stored patterns at each update step, resulting in a computational
complexity of $O(NP)$ per step. This contrasts with the $O(N^2)$
complexity of traditional Hopfield networks and can become a
bottleneck in systems with a very large number of stored patterns
($P \gg N$). Exploring efficient kernel approximation methods to
mitigate this recall complexity remains an important direction for
future work.

\subsection{Scaling Analysis of Storage Capacity and Dynamics}

\begin{figure}[t]
  \begin{center}
    \includegraphics[width=0.99\hsize]{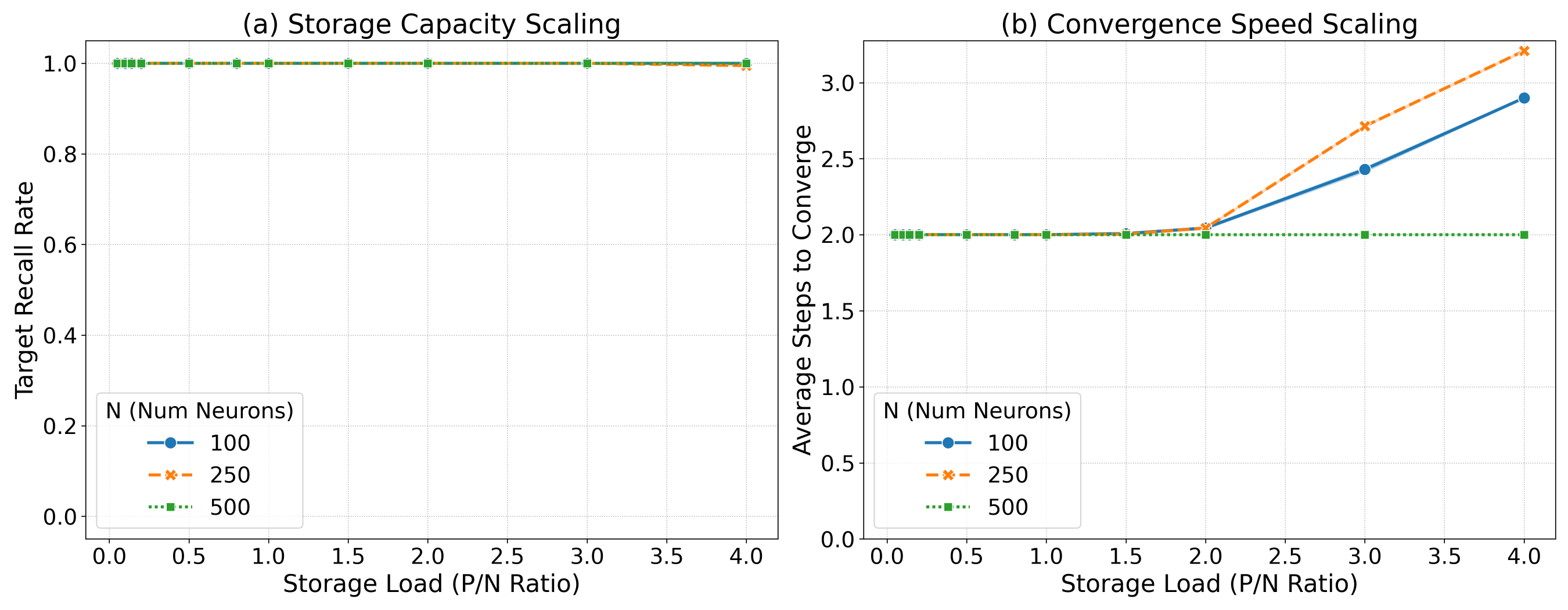}
  \end{center}
  \caption{Scaling analysis of storage capacity and convergence
    dynamics under N-dependent optimal kernel widths
    ($\gamma_{\text{opt}}$). The analysis was performed under a
    moderate noise condition (Initial Similarity = 0.8) for three
    different network sizes: $N=100, 250$, and $500$.  (a)
    \textbf{Storage Capacity Scaling:} The \textit{Target Recall Rate}
    curves for all three network sizes largely collapse onto a single
    curve, particularly in the high-performance regime
    ($P/N \lesssim 3.5$). While minor deviations are observed at the
    highest loads, this trend provides strong evidence that the
    storage capacity $P$ scales linearly with the network size
    $N$. (b) \textbf{Convergence Speed Scaling:} In contrast, the
    convergence dynamics exhibit a clear dependence on the absolute
    network size. For a given $P/N$ ratio in the high-load regime,
    larger networks converge significantly faster. This suggests that
    while the stability of attractors is governed by relative load,
    the efficiency of the recall dynamics is enhanced in
    higher-dimensional state spaces.}
\label{fig:scaling_analysis}
\end{figure}

A critical question for any associative memory model is how its
storage capacity scales with the network size, $N$. To address this
and verify that our findings are not artifacts of a specific network
size, we conducted a scaling analysis for networks of $N = 100, 250$,
and $500$ neurons.

During our initial investigation, a naive application of the
conventional kernel scaling rule, $\gamma = 1/N$, led to the
counter-intuitive result that the storage capacity coefficient $P/N$
decreased for larger networks. This motivated a focused analysis of
the interplay between $N$ and $\gamma$. We found that the optimal
kernel width, $\gamma_{\text{opt}}$, which maximizes recall
performance at high loads, exhibits a non-trivial dependence on the
network size. Specifically, our empirical analysis revealed that
larger networks require a more localized kernel to achieve maximum
capacity, with the optimal scaling factor
$c_{\text{opt}} = \gamma_{\text{opt}} N$ increasing with $N$ (e.g.,
$c_{\text{opt}} \approx 2$ for $N \leq 250$ and
$c_{\text{opt}} \approx 5$ for $N = 500$). A detailed account of this
hyperparameter optimization is provided in Section 4.4.

Armed with this crucial insight, we re-evaluated the scaling law using
the empirically determined, $N$-dependent optimal gamma values for
each network size. The results, performed under a moderate noise
condition (Initial Similarity = 0.8), are presented in Fig.~3.

As shown in Fig. 3(a), the Target Recall Rate curves for all three
network sizes now largely collapse onto a single curve, particularly
in the high-performance regime.  While minor deviations are observed
at very high loads (e.g., $P/N = 4.0$), the performance is
predominantly determined by the relative storage load $P/N$,
irrespective of the absolute network size $N$. This result provides
strong evidence that, under appropriately scaled kernel parameters,
the storage capacity of the KLR-trained network follows the
fundamental linear scaling law, $P \propto N$. This confirms that the
high capacity coefficients reported are a general property of the
model and not a finite-size effect.

Interestingly, while the static property of storage capacity follows a
simple $P/N$ scaling, the temporal dynamics of the recall process
exhibit a more complex, scale-dependent behavior. Fig. 3(b) plots the
average number of steps to convergence. In contrast to the performance
curves, the convergence speed does not collapse onto a single
curve. For any given $P/N$ ratio in the high-load regime, larger
networks converge significantly faster than smaller ones. For example,
at $P/N = 4.0$, the $N=500$ network converges in approximately 2
steps, whereas the $N=100$ network requires nearly 3 steps.

This dissociation between the scaling of capacity and the scaling of
convergence speed is a non-trivial finding. It suggests that while the
existence and size of the basins of attraction are governed by the
relative density of patterns ($P/N$), the efficiency of the trajectory
towards the attractor's center is enhanced in higher-dimensional state
spaces. This intriguing scale-dependent acceleration of dynamics will
be further explored in the Discussion section.

\subsection{Hyperparameter Sensitivity}

\begin{figure}[t]
  \begin{center}
    \includegraphics[width=0.99\hsize]{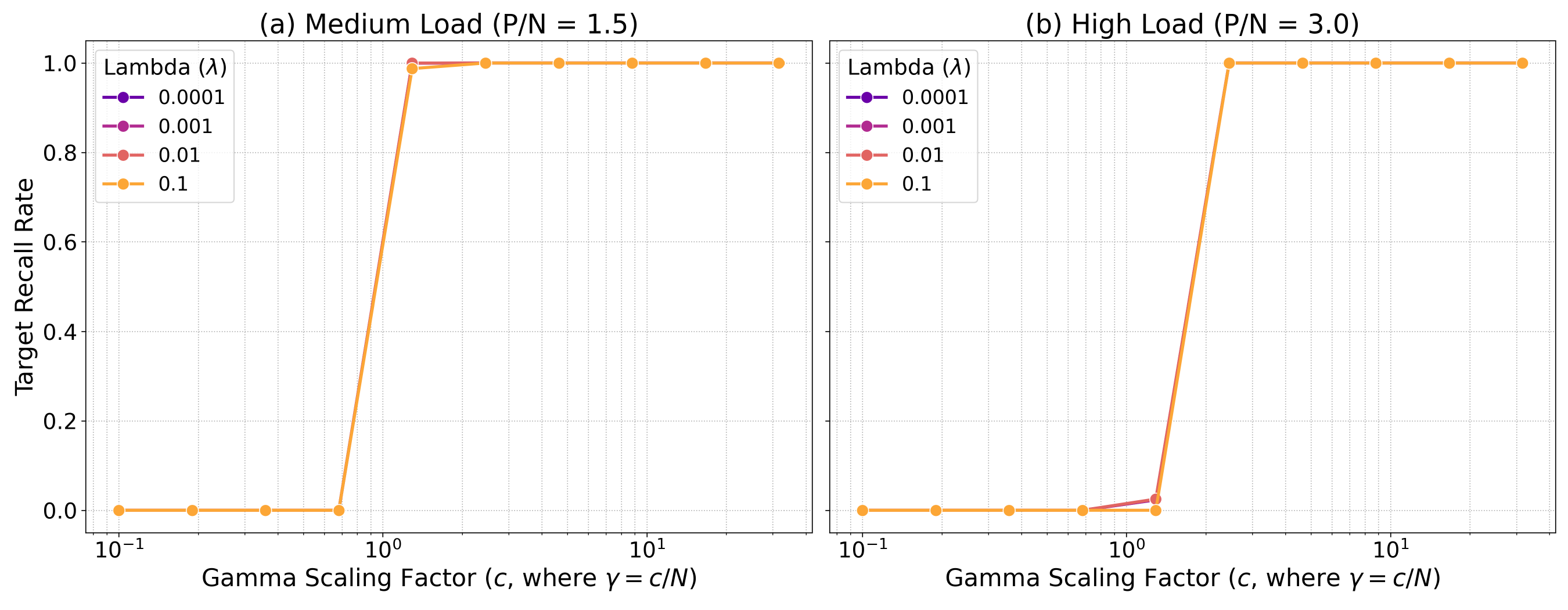}
  \end{center}
  \caption{Hyperparameter sensitivity analysis for a network with
    $N=100$ neurons.  The plots show the Target Recall Rate as a
    function of the scaling factor $c$ (where $\gamma = c/N$), for
    four different values of the L2-regularization parameter
    $\lambda$.  All data points are averaged over five master seeds
    under a moderate noise condition (Initial Similarity = 0.8).  (a)
    Performance at a medium storage load ($P/N$ = 1.5).  (b)
    Performance at a high storage load ($P/N$ = 3.0).  The results
    reveal a sharp phase transition, indicating that a minimum level
    of kernel locality ($c \gtrsim 1$) is required for successful
    recall.  Crucially, the performance is remarkably robust to the
    choice of $\lambda$ over several orders of magnitude, as the
    curves for different $\lambda$ values are nearly
    indistinguishable.}
\label{fig:sensitivity_analysis}
\end{figure}

Finally, we investigated the sensitivity of the network's performance
to its two key hyperparameters: the kernel width $\gamma$
(parameterized by the scaling factor $c = \gamma N$) and the
L2-regularization parameter $\lambda$. A robust performance across a
reasonable range of these parameters is crucial for the practical
applicability of the model. We performed a systematic sweep of $c$ and
$\lambda$ for a network of $N=100$ under both medium ($P/N = 1.5$) and
high ($P/N = 3.0$) storage loads.

The results are presented in Fig.~4. The plots reveal two fundamental
characteristics of the KLR-trained network. First, the network's
performance exhibits a sharp phase transition with respect to the
kernel locality $c$. For both medium and high load conditions, the
\textit{Target Recall Rate} is near zero for small $c$ values
($c < 1$) and abruptly jumps to near-perfect recall once $c$ crosses a
critical threshold. This indicates that a minimum level of kernel
locality is essential for the learning algorithm to successfully form
distinct, stable attractors. Interestingly, the critical threshold for
$c$ appears to be slightly larger for the high-load case (Fig. 4(b))
than for the medium-load case (Fig. 4(a)), corroborating our earlier
finding that higher memory congestion requires a more localized kernel
to mitigate inter-pattern interference. Second, and most strikingly,
the network's performance is remarkably robust to the choice of the
regularization parameter $\lambda$. As shown in both plots, the
performance curves for four different $\lambda$ values, spanning three
orders of magnitude (from 0.0001 to 0.1), are nearly indistinguishable
from one another. This result is of significant practical importance,
as it suggests that the KLR Hopfield network does not require
meticulous fine-tuning of the regularization parameter to achieve high
performance. The learning mechanism appears to be inherently stable
and capable of finding effective solutions across a very broad range
of $\lambda$.

In summary, this sensitivity analysis demonstrates that while the
kernel width $\gamma$ is a critical parameter that must be chosen to
ensure sufficient kernel locality (i.e., $c = \gamma N$ must be large
enough), the model is exceptionally robust with respect to the
regularization parameter $\lambda$. This provides a clear and
practical guideline for the design and application of KLR-trained
associative memories.

\section{Discussion}
\label{sec:discussion}
Our extensive quantitative analysis has provided a comprehensive
picture of the attractor landscape in KLR-trained Hopfield networks. A
central finding is the emergence of a remarkably ``clean'' landscape,
characterized by a near-zero rate of spurious fixed points,
particularly under high storage loads with moderate noise
(Fig.~1(b)). This strong suppression of spurious states is a key
factor behind the model's high fidelity. But what is the mechanism
behind this suppression?

We hypothesize that two aspects of KLR learning are crucial. First,
the kernel trick implicitly maps the input patterns into a
high-dimensional feature space where they become more easily
separable~\cite{Scholkopf2001}. This enhanced separability allows the
learning algorithm to find decision boundaries that create wide,
well-defined basins of attraction for the stored patterns, effectively
pushing potential spurious minima away from the recall
trajectories. Second, the logistic loss function, combined with
L2-regularization, encourages the formation of a ``smooth'' energy
landscape. This property is akin to the margin maximization in SVMs;
by pushing the decision boundaries away from the data points, the
learning process effectively deepens the basins of attraction around
the stored patterns.  Unlike Hebbian learning, which is based on
simple pairwise correlations and can create many noisy local minima,
the KLR objective seeks a solution with a large margin, which tends to
flatten the landscape in regions far from the decision
boundaries. This landscape smoothing further inhibits the formation of
unintended spurious attractors. A detailed theoretical investigation
of this mechanism is the subject of our companion
paper~\cite{tamamori2025c}.

\textbf{Generality of High-Capacity Kernel-Based Memories}: Our direct
comparison between KLR and KRR (Fig. 2) yields a crucial insight: the
remarkable ability to form a clean, high-capacity attractor landscape
is not a feature unique to the logistic loss function of KLR. The
qualitatively similar performance of KRR suggests that this is a more
general property of kernel regression methods applied to Hopfield-type
networks. The key mechanism appears to be the kernel trick itself,
which maps patterns into a high-dimensional feature space where they
become more easily separable, allowing the learning algorithm to
sculpt well-defined attractor basins.

Theoretically, this similarity is further supported by the connection
between the two algorithms. It is well-established that the
optimization of the KLR objective via Newton's method (Iteratively
Reweighted Least Squares) is mathematically equivalent to solving a
sequence of weighted Kernel Ridge Regression
problems~\cite{Bishop2006, GPML2006, Murphy2022}. In this view, KLR is
an iteratively refined version of KRR that focuses on difficult
patterns. The fact that standard KRR performs comparably suggests that
for the random patterns used in this study, such iterative refinement
yields only marginal gains. While their performance is similar, the
significant computational advantage of KRR's non-iterative solution
makes it a compelling practical alternative, highlighting a clear
trade-off between algorithmic approaches.

\textbf{The Non-Trivial Scaling of Stability}: The scaling analysis
(Fig. 3) revealed one of the most profound findings of this study: the
relationship between network size $N$ and the optimal kernel width
$\gamma$ is non-trivial. Contrary to the conventional $\gamma = 1/N$
rule of thumb, we found that maintaining optimal storage capacity
requires $\gamma$ to be scaled such that the factor $c = \gamma N$
increases with $N$. This indicates that larger networks necessitate
more localized kernels to effectively manage the increasingly complex
web of inter-pattern interactions in a high-dimensional state
space. In essence, as the memory space becomes more crowded with
patterns in a higher-dimensional state space, each stored memory must
become more ``introverted,'' narrowing its field of influence to avoid
interfering with its neighbors. This scale-dependent requirement for
locality is a new, fundamental principle for designing high-capacity
kernel-based associative memories. The requirement for locality can
also be interpreted as a means to maintain the effective rank of the
kernel matrix high enough to ensure linear separability in the
high-dimensional feature space. As a practical guideline derived from
our findings, optimal performance for random patterns can be achieved
by selecting $\gamma$ such that the scaling factor $c = \gamma N$ is
in the range of approximately 2 to 5.  Once this principle is
respected, we confirmed that the storage capacity follows the expected
linear scaling law, $P \propto N$, demonstrating the model's soundness
as a scalable architecture.  However, we note that the observed linear
scaling law has been verified only for network sizes up to $N=500$,
and its validity for substantially larger systems remains
uncertain. Moreover, practical scalability is ultimately constrained
by the $O(NP)$ computational cost of recall updates. Clarifying how
these factors interact in larger regimes remains an important
direction for future investigation.

\textbf{Dissociation of Performance and Dynamics Scaling}:
Intriguingly, while storage capacity (a static property) scales with
the relative load $P/N$, the recall dynamics (a temporal property) do
not. As shown in Fig. 3(b), for a given $P/N$, larger networks
converge significantly faster. This dissociation suggests that the
geometry of the attractor landscape changes with $N$ in a subtle
way. We hypothesize that in higher-dimensional state spaces, the
energy landscape becomes ``smoother'' or less rugged, allowing the
state to follow a more direct path to the attractor minimum. This
``blessing of dimensionality'' for dynamics, coexisting with a simple
$P/N$ scaling for capacity, is a fascinating aspect of this system
that warrants further theoretical investigation.

\textbf{Robustness and Practical Implications}: Our hyperparameter
sensitivity analysis (Fig. 4) provided both a theoretical insight and
a practical guideline. The sharp phase transition with respect to the
kernel locality $c$ underscores its criticality: a minimum level of
locality is non-negotiable for stability. Conversely, the astonishing
robustness of performance to the regularization parameter $\lambda$
over several orders of magnitude is a highly desirable feature. It
implies that, once the kernel is appropriately scaled, the system is
remarkably stable and does not require meticulous fine-tuning, greatly
enhancing its practical applicability.

One limitation of this work is that the stored patterns were
restricted to independent random binary vectors. While this controlled
setting enables precise quantitative analysis of attractor statistics,
real-world data, such as images or time series, typically exhibit
strong statistical structure. Evaluating the proposed framework on
such structured datasets therefore remains an important direction for
future work. Such evaluations would help clarify the extent to which
the observed scaling laws and attractor properties generalize beyond
random pattern ensembles.

In conclusion, this work has moved the understanding of KLR-trained
Hopfield networks from a collection of performance benchmarks to a
well-characterized dynamical system. By revealing the generality of
the approach (via KRR comparison), the non-trivial scaling laws of its
parameters, and the robust nature of its performance, we have
established a solid foundation for both the theoretical exploration
and practical engineering of the next generation of high-capacity,
brain-inspired associative memories.

\section{Conclusion}
\label{sec:conclusion}
In this paper, we presented a comprehensive quantitative analysis of
the attractor landscape in Hopfield networks trained with Kernel
Logistic Regression, addressing critical questions regarding the
model's performance, generality, and scalability. Through a series of
rigorous, statistically validated simulations, we have moved beyond
the initial discovery of high capacity to establish a deeper, more
principled understanding of this powerful associative memory system.

Our findings confirm that KLR training sculpts a remarkably clean and
robust attractor landscape, characterized by a near-zero rate of
spurious states across a broad range of parameters and exceptionally
fast recall dynamics. Our comparative analysis revealed that this high
performance is a general feature of kernel regression methods, though
KLR and KRR present a trade-off between iterative flexibility and
closed-form efficiency.

Most significantly, we uncovered a non-trivial, scale-dependent
scaling law for the kernel width, demonstrating that optimal storage
capacity is intrinsically linked to the appropriate tuning of the
kernel's locality with network size. Under this optimized scaling, we
provided definitive evidence that the storage capacity scales linearly
with network size, $P \propto N$, solidifying the model's foundation
as a scalable architecture. Furthermore, we demonstrated the model's
practical robustness, showing its high performance to be remarkably
insensitive to the choice of the regularization parameter.

From a practical standpoint, a remaining challenge of kernel-based
recall is the $O(NP)$ computational cost per update step, which may
become prohibitive when the number of stored patterns is
large. Potential approaches to mitigating this issue include kernel
approximation techniques (e.g., random feature expansions), prototype
selection methods, and sparsification of the dual
coefficients. Exploring such approaches while preserving the attractor
structure remains an important direction for future work.

In summary, this work provides a clear set of empirical findings and
practical design principles for constructing high-capacity,
kernel-based Hopfield networks. By systematically characterizing the
properties of the attractor landscape and its dependence on key
parameters, we have laid a solid foundation for future theoretical
work and explorations into alternative loss functions, such as hinge
loss for sparsity-induced efficiency. This establishes a robust
framework for the engineering of next-generation, brain-inspired
memory systems.

\funding
Not applicable.

\conflictsofinterest
The author declares no competing interests.

\authorcontribution
The sole author contributed to the present work.

\aitools
The author used ChatGPT (GPT-5.2) and Gemini 2.5 Pro for proofreading
the English manuscript.

\appendix
\section{Convergence of KLR Learning}
\label{sec:appendix_convergence}

\begin{figure}[h]
    \centering
    \includegraphics[width=0.7\textwidth]{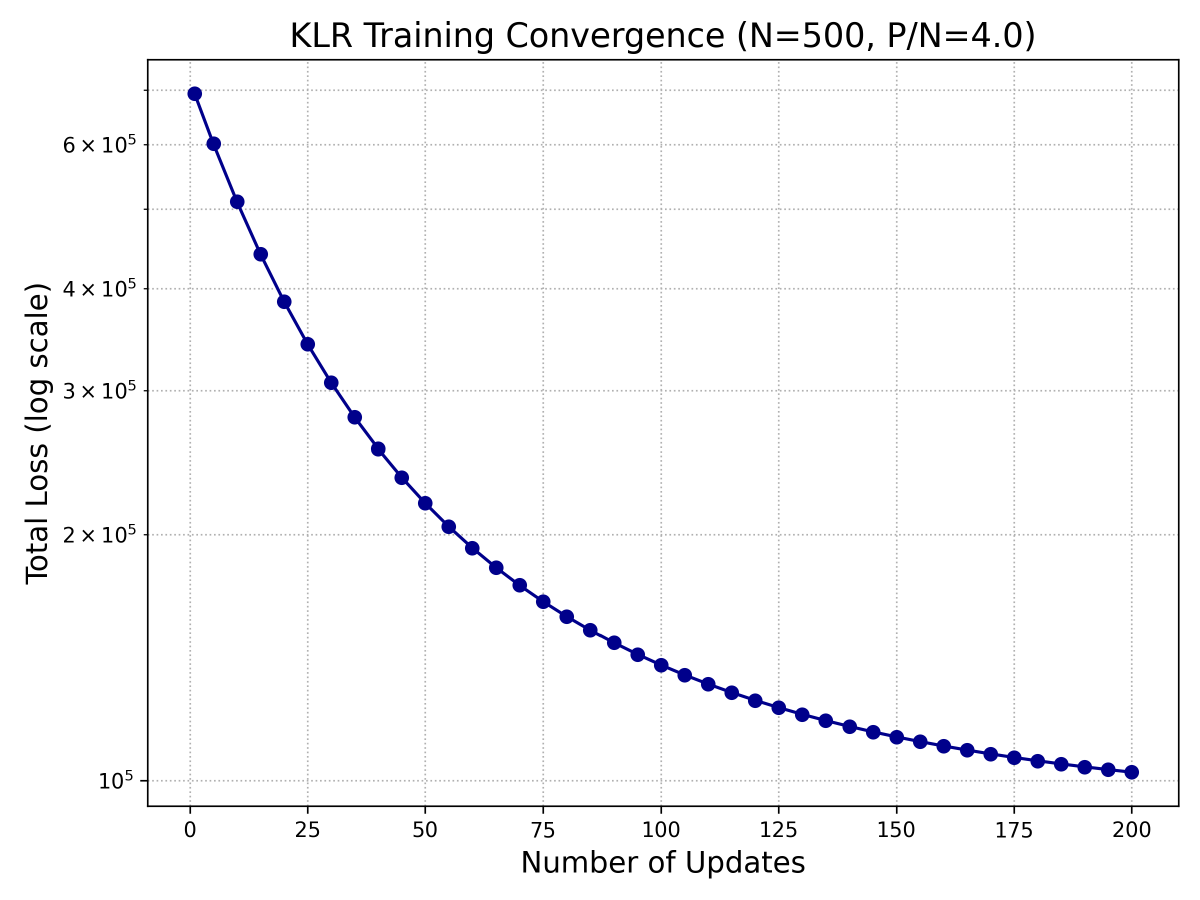}
    \caption{Training convergence of Kernel Logistic Regression
      (KLR). The plot shows the total loss as a function of the number
      of updates for a network under a high-load condition ($N=500$,
      $P/N=4.0$). The curve demonstrates that the loss effectively
      converges within the 200 updates used in our experiments.}
    \label{fig:learning_curve}
\end{figure}

To validate that the number of updates used for Kernel Logistic
Regression (KLR) training was sufficient for the convergence of the
model parameters, we analyzed the learning curve under a challenging,
high-load condition.  Fig.~\ref{fig:learning_curve} plots the total
loss (negative log-likelihood plus L2-regularization term) as a
function of the number of gradient descent updates for a network with
$N=500$ neurons and a storage load of $P/N=4.0$.

As the figure illustrates, the loss decreases by several orders of
magnitude within the initial 150 updates and visibly approaches a
plateau. The rate of decrease becomes negligible towards the end of
the 200-update training period. This confirms that 200 updates are
adequate to achieve a well-converged model, ensuring that the
performance results reported in the main text are not limited by
incomplete training.

\section{Validation of Attractor Classification}
\label{sec:appendix_Attractor_Classification}
\begin{figure}[h]
    \centering
    \includegraphics[width=0.75\textwidth]{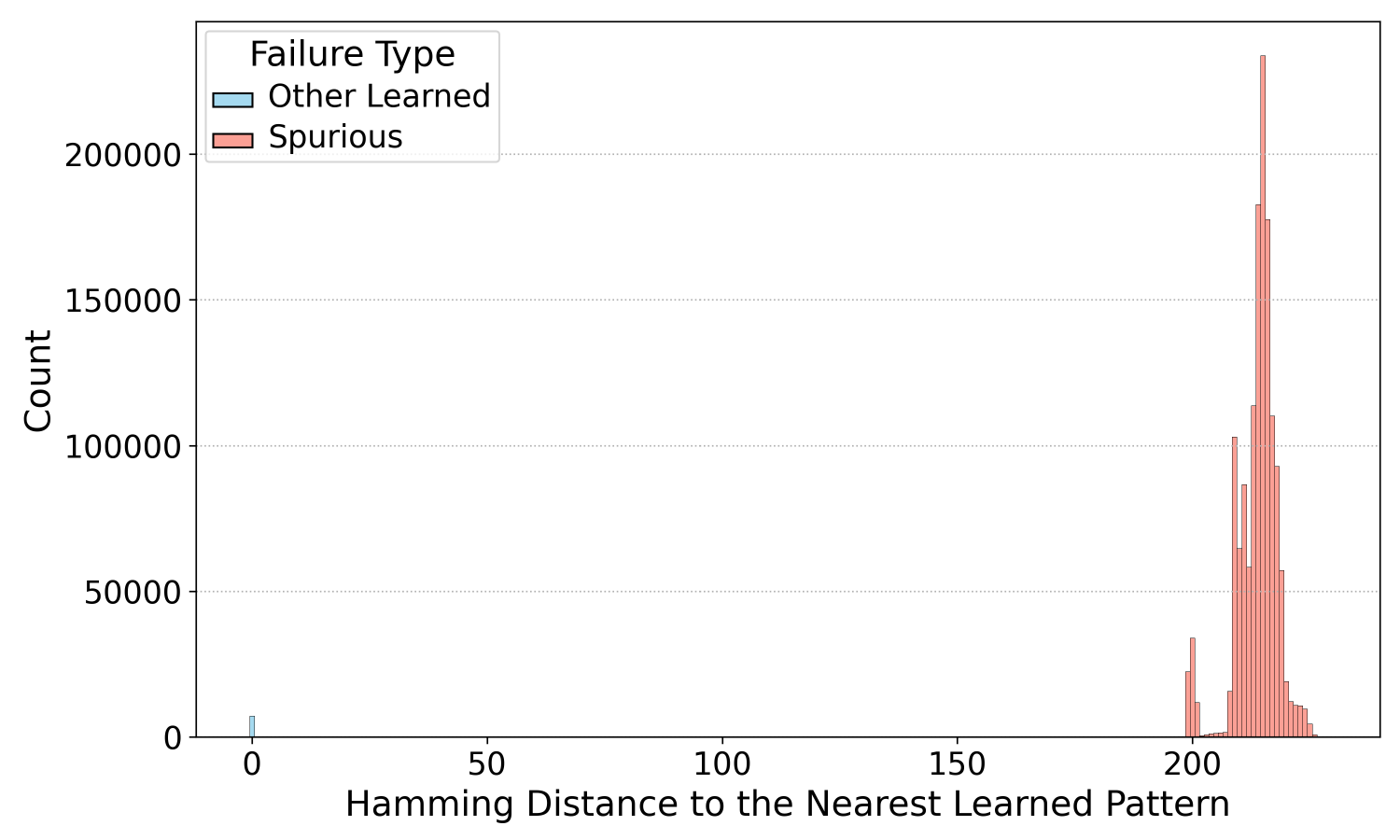}
    \caption{Histogram of the Hamming distance from the final state of
      a failed trial to the nearest learned pattern. The distribution
      is shown for trials classified as Other Learned (blue) and
      Spurious (red). The perfect separation of the two distributions
      validates our strict, equality-based classification criterion.}
    \label{fig:hamming_distance}
\end{figure}

To validate the strict equality-based classification of attractors
used in our study, we analyzed the distribution of Hamming distances
between final states of failed recall trials and their nearest learned
patterns. Fig.~\ref{fig:hamming_distance} shows the histogram of
these distances, separated by failure type (Other Learned
vs. Spurious).

As expected, all trials classified as Other Learned resulted in final
states with a Hamming distance of exactly 0 to a non-target learned
pattern. Conversely, all trials classified as Spurious resulted in
final states with a significant, non-zero Hamming distance to any of
the learned patterns. The clear separation between these two
distributions, with no spurious attractors found in close proximity
(e.g., Hamming distance $< 10$) to any learned pattern, confirms the
validity of our classification methodology.

\section{Analysis of Non-Stationary Dynamics}
\label{sec:appendix_dynamics}
\begin{figure*}[h]
    \centering
    \begin{minipage}[b]{0.49\textwidth}
        \centering
        \includegraphics[width=\linewidth]{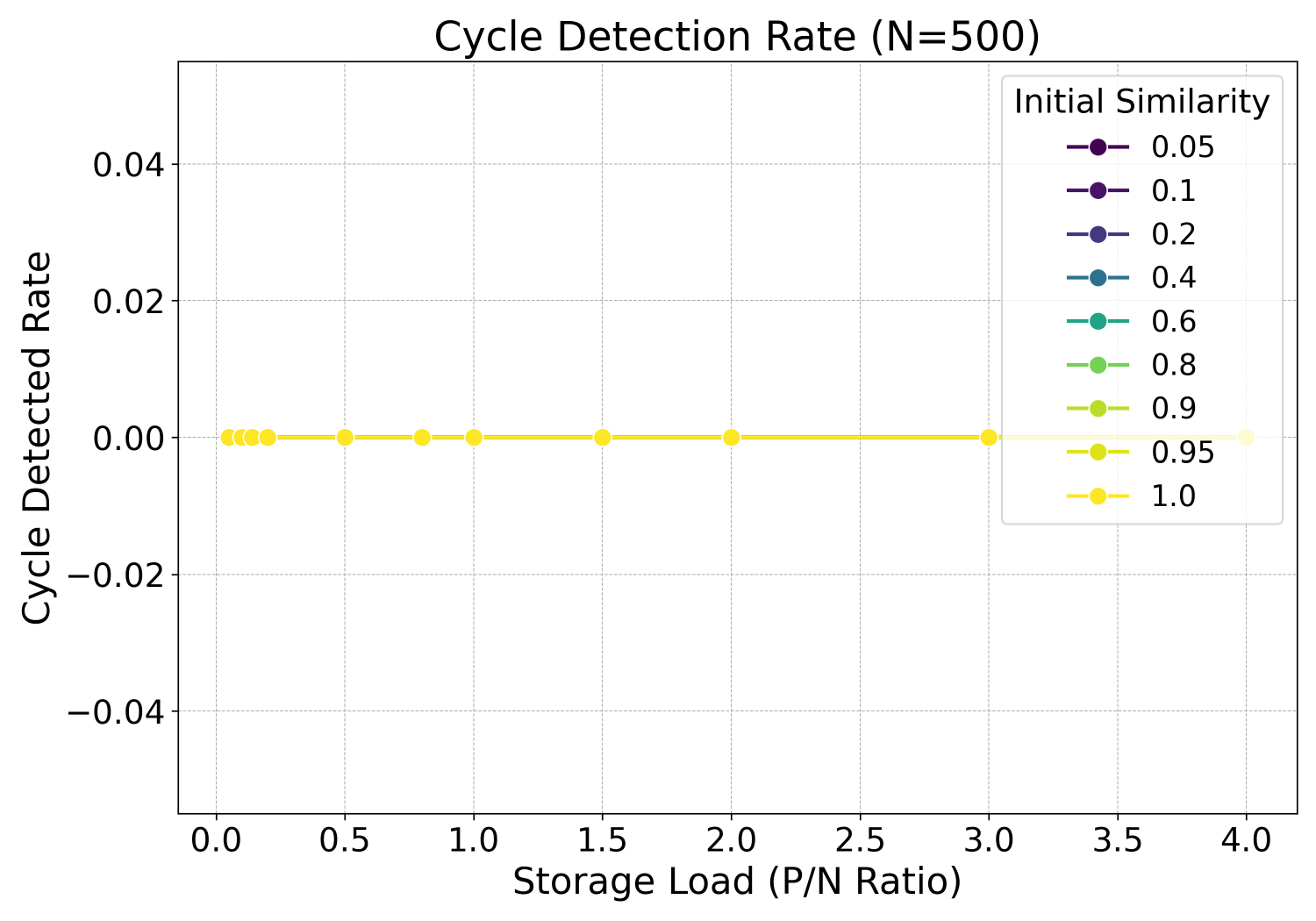}
        \centerline{(a) Cycle Detection Rate vs. Storage Load}
        \label{fig:capacity}
    \end{minipage}
    \hfill
    \begin{minipage}[b]{0.49\textwidth}
        \centering
        \includegraphics[width=\linewidth]{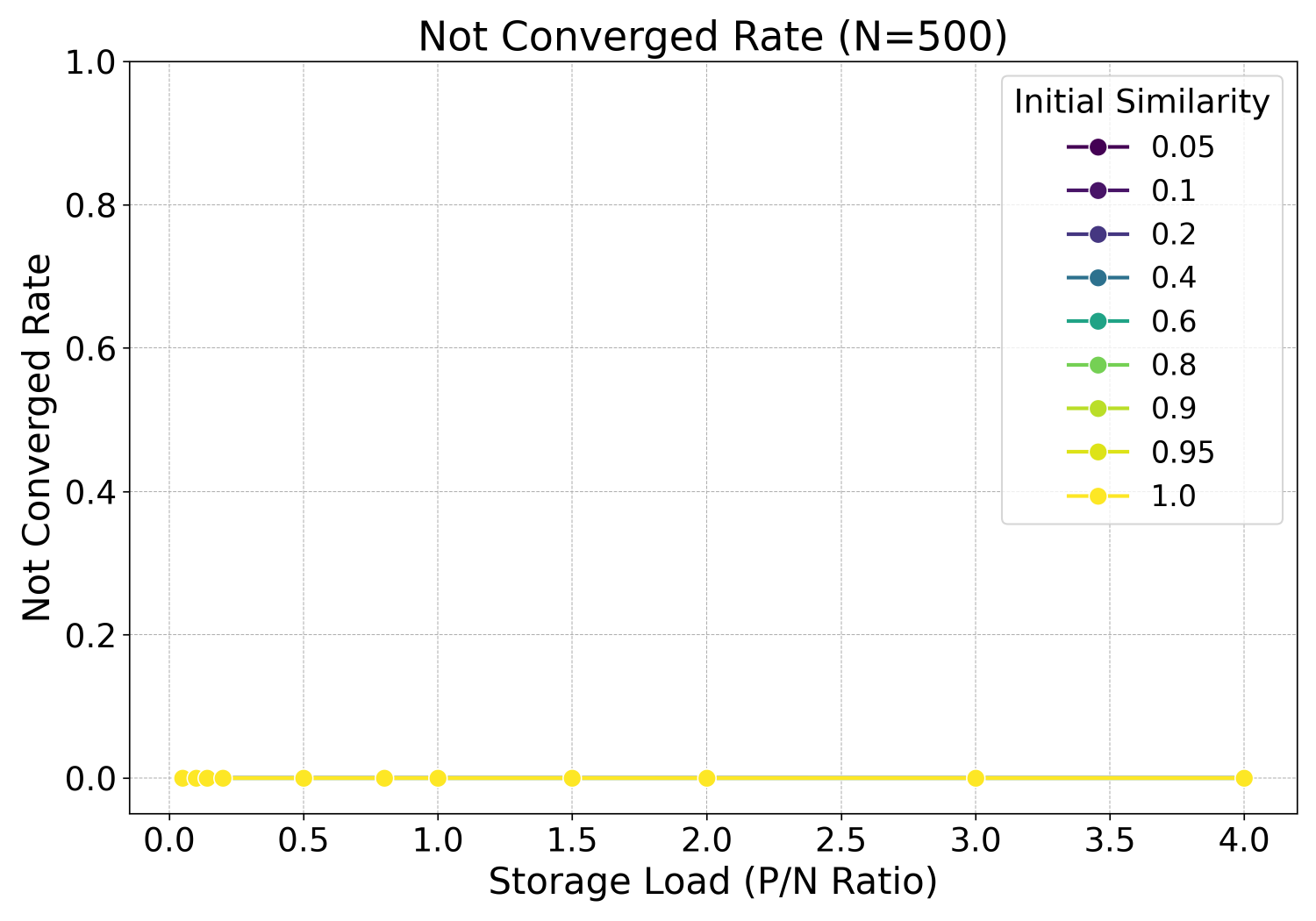}
        \centerline{(b) Not Converged Rate vs. Storage Load}
        \label{fig:error_decomp}
    \end{minipage}
    \caption{Analysis of non-stationary dynamics for $N=500$. All data
      points represent the mean over five master seeds. Shaded regions
      indicate 95\% confidence intervals, though they are mostly
      invisible as the rates are zero.  (a) The proportion of trials
      that converged to a limit cycle.  (b) The proportion of trials
      that did not converge within 30 steps.  Both rates are
      effectively zero across all conditions, confirming that the
      network dynamics are dominated by convergence to fixed points.}
    \label{fig:appendix_dynamics}
\end{figure*}

To ensure that our analysis, which primarily focuses on fixed point
attractors, is not biased by the presence of other dynamical
behaviors, we systematically quantified the rate of convergence to
non-stationary states. This appendix presents the results for the
rates of convergence to limit cycles and non-convergence within the
maximum simulation steps.

Figure \ref{fig:appendix_dynamics} shows the \textit{Cycle Rate} and
the \textit{Not Converged Rate} as a function of storage load ($P/N$)
for a network of $N=500$. As can be seen in
Fig. \ref{fig:appendix_dynamics}(a), the \textit{Cycle Rate} is
effectively zero across all tested conditions, with no trials
converging to a limit cycle with a period of 2 or more. Similarly,
Fig. \ref{fig:appendix_dynamics}(b) shows that the \textit{Not
  Converged Rate} is also zero, confirming that every recall trial
reached a stable state (either a fixed point or a cycle) within the 30
update steps.

These results validate our focus on fixed point attractors, as they
overwhelmingly dominate the dynamics of the KLR-trained network under
the conditions studied in this paper.

\end{document}